# Enhancing the Accuracy of Biometric Feature Extraction Fusion Using Gabor Filter and Mahalanobis Distance Algorithm

[1]Ayodeji S. Makinde, [2]Yaw Nkansah-Gyekye, [3]Loserian S. Laizer

[1,2,3] School of Computational and Communication Science and Engineering, NM-AIST, Tanzania

[1] makindea@nm-aist.ac.tz, [2] yaw.nkansah-gyekye@nm-aist.ac.tz, [3] loserian.laizer@nm-aist.ac.tz

*Abstract*- **Biometric recognition systems have advanced significantly in the last decade and their use in specific applications will increase in the near future. The ability to conduct meaningful comparisons and assessments will be crucial to successful deployment and increasing biometric adoption. The best modality used as unimodal biometric systems are unable to fully address the problem of higher recognition rate. Multimodal biometric systems are able to mitigate some of the limitations encountered in unimodal biometric systems, such as non-universality, distinctiveness, non-acceptability, noisy sensor data, spoof attacks, and performance.**

**More reliable recognition accuracy and performance are achievable as different modalities were being combined together and different algorithms or techniques were being used. The work presented in this paper focuses on a bimodal biometric system using face and fingerprint. An image enhancement technique (histogram equalization) is used to enhance the face and fingerprint images. Salient features of the face and fingerprint were extracted using the Gabor filter technique. A dimensionality reduction technique was carried out on both images extracted features using a principal component analysis technique. A feature level fusion algorithm (Mahalanobis distance technique) is used to combine each unimodal feature together. The performance of the proposed approach is validated and is effective.**

I. INTRODUCTION

*A.    Background*

With the advancement in networking, communication, and mobility in today's electronically wired information society, the need for accurate and reliable feature extraction of biometric traits in multibiometric systems is very crucial. Feature extraction refers to the process of generating a compact but expressive digital representation of the underlying biometric trait, called a template which contains only the salient discriminatory information that is essential for recognizing the person [2][4][10][24].

A good biometric is characterized by the use of features that are highly unique, so that the chance of any two people having the same characteristics will be minimal, stable, does not change over time, easily captured in order to provide convenience to the user, and prevent misrepresentation of the feature [19]. For multi-biometric recognition to have low False Rejection Rate (FRR) and False Acceptance Rate (FAR), an efficient feature extraction algorithm is needed.

In order to provide accurate recognition of individuals, the most discriminating information present in a bimodal face and fingerprint system must be extracted so that comparison between templates can be made. In this paper two prominent modalities (face and fingerprint images) are used for the bimodal features extraction. Gabor filter is used in extracting the salient feature from the two modalities, due to its ability to extract maximum information from local image regions, and being invariant against translation, rotation, and variations [1][7][8]. Although there are other algorithms which also perform better with different characteristics in feature extraction of face and fingerprint images, each of these algorithms cannot be used in extracting both face and fingerprint features at the same time. [6] used Gabor filter for the extraction of fingerprint features with an accuracy of 97.2%. Likewise, [13] also used Gabor filter for fingerprint feature extraction.

In the recent years Gabor filter was noted for the extraction of face images [1][28] than that of the fingerprint images. The problem in multimodal biometric fusion is that each modality was extracted with different algorithm which makes it difficult to fuse together [2][3]. However, based on empirical literature, it was noted that Gabor filter has a good feature extraction accuracy on both images especially at the feature extraction level.

*B. Biometric system*

Biometric systems are technologies used to identify or recognize individuals based on their physiological or behavioral characteristics. They capture biometric features of a person and extract a set of salient features that are compared with a set of features in the form of a template already extracted and stored in the database of the same person. Several human characteristics that can be used as the basis for biometric systems include a person's face, fingerprint, iris, DNA. A biometric-based authentication system consists of two main phases, namely, enrollment and recognition [2][3].

*C. Why multimodal biometric system?*

Unimodal biometric systems are affected by the following problems [15][16]:

- Noisy sensor data: The sensed data might be noisy or distorted.
- Non-universality: While every user is expected to possess the biometric trait being acquired, in reality, it is possible for a subset of the users to be unable to provide a particular biometric.
- Lack of individuality: While a biometric trait is expected to vary significantly across individuals, there may be large similarities in the feature sets used to represent these traits. This limitation restricts the discriminality provided by the biometric trait.
- Spoof attacks: An impostor may attempt to spoof the biometric trait of a legitimate enrolled user in order to circumvent the system.

Due to these practical problems, the error rates associated with unimodal biometric systems are quite high which make them unacceptable for deployment in critical security applications. Some of the problems that affect unimodal biometric systems can be overcome by using multimodal biometric systems. Fusing two or more traits together tends to address some of the problems being faced in the unimodal biometric systems. In this paper, face and fingerprint images were fused together for biometric recognition due to the fact that they have some advantages over other modalities such as availability, collectability, mostly country or religion conflict free and long existence. Although there are other modalities such as iris, retina, and vein pattern with high uniqueness, universality, permanence, performance, resistance to circumvention, they are difficult to maintain. Thus, it makes face and fingerprint modalities to be some of the most researched and mature fields of authentication. Figure 1 illustrates different fusion of biometrics and various fusion scenarios that are feasible in multimodal biometric systems.

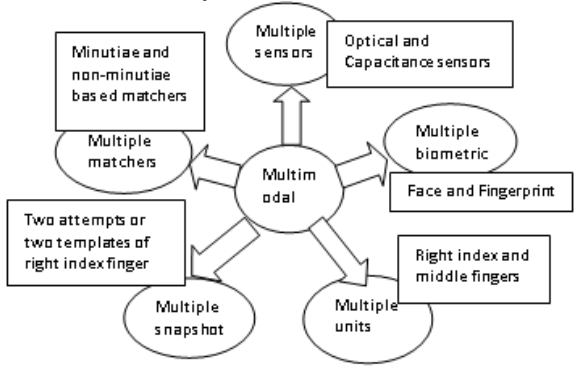

Figure 1: Scenarios in a multimodal biometric system [20]

This paper proposes an enhanced biometric feature extraction fusion using Gabor filter and Mahalanobis distance. The next section presents the literature review. Section 3 presents the material and method used. Results and discussion are given in Section 4. Conclusion and recommendations are made in the last section.

*D. Originality and contribution*

In this paper, we present a novel way of extracting the features of face and fingerprint modalities which form a bimodal biometric system using a Gabor filter. The extraction rate obtained from the proposed method was compared with other forms of feature extraction algorithms used in both face and fingerprint. The contribution of this paper was to enhance the accuracy and performance of multimodal biometric systems at the feature extraction level. One of the challenges of multimodal biometric systems is the difficulty of fusing two modalities of different feature extraction algorithm at the feature extraction level.

II. LITERATURE REVIEW

In this section, we review the different biometric feature extraction algorithms that have been used for both face and fingerprint recognition.

*A. Face feature extraction algorithm review*

[12] presented an automatic facial feature extraction method using genetic algorithm. The method was based on the edge density distribution of the image. In the preprocessing stage a face is approximated to an ellipse and a genetic algorithm is applied to search for the best ellipse region match in which genetic algorithm is applied to extract the facial features. The experimental results validate that the proposed method is capable of automatically extracting features from various video images effectively under natural lighting environments and in the presence of a certain amount of artificial noise and of multiface oriented with angles. They proposed that the algorithm can be improved so that it can be applied to real world problem, by incorporating more characteristics in the fitness function during the evolutionary process and also to extract features on faces containing either beard or mustache.

[1] presented a neural network system for face recognition in which Gabor filter was used for extracting the salient features, and the feature vectors which were based on Fourier Gabor filters is used as input of their classifier which is a Back Propagation Neural Network (BPNN). Also, due to the large dimension of the input vector, a dimensionality reduction algorithm called Random Projection (RP) was used. Their experimental results demonstrate that using Gabor filter for face feature extraction proves the robustness of their solution, due to the fact that more salient features were being extracted.

[20] used independent component analysis (ICA) method for the feature extraction of face images for face recognition. They used a version of ICA derived from the principle of optimal information transfer through sigmoidal neurons. ICA was performed on face images in the FERET database under two different architectures, one of which treated the images as random variables and the pixels as outcomes, and a second one treated the pixels as random variables and the images as outcomes. The first architecture found spatially local basis images for the faces. The second architecture produced a factorial face code. The experimental results express that both ICA representatives were superior to representation based on PCA for recognizing faces across days and changes in expression. Also, a classifier that combined the two ICA representations gave the best performance.

[22] presented in their paper, a new model bidirectional associative memory (BAM) inspired architecture that can ultimately create its own set of perceptual features. The resulting model inherits properties such as attractor-like behavior and successful processing of noisy inputs, while being able to achieve principal component analysis (PCA) tasks such as feature extraction and dimensionality reduction. The model is tested by simulating image reconstruction and blind source separation tasks. Simulations show that the model fares particularly well compared to existing neural PCA and independent component analysis (ICA) algorithms. It is argued that the model possesses more cognitive explanative power than any other nonlinear/linear PCA and ICA algorithm.

[21] stated that LDA is one of the most commonly used techniques for data classification and dimensionality reductions. They further explain that LDA easily handles the situation where the within-class frequencies are unequal and their performance has been examined on randomly generated test data. This method maximizes the rate of between-class variance to the within-class variance in any particular data set thereby guaranteeing maximal separability. The prime difference between LDA and PCA is that the PCA does more of feature classification and LDA does data classification. In PCA, the shape and location of the original data set changes when transformed into a different space, whereas LDA does not change the location, but only tries to provide more class separability and draws a decision region between the given classes.

[24] proposed a weighted 2D Principal Component Analysis (2DPCA) model which addresses some of the challenges faced in using 2DPCA as a face recognition extractor. [27] make use of 2DPCA in the feature extraction of face recognition in which face images were represented in matrices or 2D images form compared to the conventional PCA which represents images in vector form. Also, they stated that using 2D images directly is quite simple and local information of the original images is preserved appropriately, which may bring more important features for facial representation. With all these advantages, not all the face images are easy to recognize. For example, frontal face images are easier to be recognized than profile face images, which subsequently led to the proposed weighted-2DPCA model to deal with some practical situations in which some face images in database are difficult due to their poses (front of the profile) or their qualities (noise, blur).

The algorithm used in a Weighted-2DPCA model for the face model construction consists of the following steps:

Step 1: Compute the mean image

$$\tilde{A} = \frac{\sum_{i=1}^{N} w_i A^{(i)}}{\sum_{i=1}^{N} w_i} \qquad (1)$$

Step 2: Compute the matrix

$$G = \frac{\sum_{i=1}^{N} w_i (A^{(i)} - \tilde{A})^T (A^{(i)} - \tilde{A})}{\sum_{i=1}^{N} w_i} \qquad (2)$$

Step 3: Compute eigenvectors $\{\Omega_1, \Omega_2, ..., \Omega_n\}$ and eigenvalues $\{\lambda_1, \lambda_2, ..., \lambda_n\}$ of $G$, where $w_i$ is the weight and $A$ is the image matric.

The results indicate that 97.3% accuracy using Weighted-2DPCA compared to 96.2% accuracy using 2DPCA and 95.2% accuracy using conventional PCA.

*B. Fingerprint feature extraction algorithm review*

[6] presented a Gabor filter based method for directly extracting fingerprint features from gray level images without the introduction of preprocessing of the original images acquired at the sensor level. Their proposed method is more suitable than conventional methods for a small scale fingerprint recognition system. Their experimental results demonstrate that the recognition to the k-nearest neighbor classifier using the proposed Gabor filter based features has an accuracy of 97.2% with 3-NN classifiers.

[14] presented a fingerprint feature extraction method through which minutiae are extracted directly from original gray-level fingerprint images without binarization and thinning. Their algorithm improves the performance of the existing ones along this stream by using the following steps: first, they preprocess the fingerprint images by making use of Gabor filter. Second, they find the computation of the orientation field which is very crucial in order to trace the ridgelines in the fingerprint images correctly. Third, they determined the starting pixels to trace ridgelines, once a starting pixel in a ridge has been decided, the minutiae are recorded. Their experimental results indicate that the approach can achieve better performance in both efficiency and robustness.

[9] presented a novel ridge tracing approach for extraction of fingerprint features directly from gray scale images. With this method, they made use of contextual information gathered

during the tracing process to better handle noisy regions. Their experimental results have been compared with other feature extraction algorithms such as Gabor based filtering as well as the original ridge tracing work, which clearly show that their proposed approach makes a ridge tracing more robust to noise and makes the extracted features more reliable.

[13] presented a set of fingerprint recognition algorithms which includes Gamma controller normalization and equalizer, fingerprint image division, fingerprint image binarization and different direction Gabor filter for feature extraction by taking into account both the global and local features of the fingerprints which were based on the fingerprint image enhancement and the texture using Gabor filter. The experimental results illustrate that the proposed algorithm can avoid all sorts of false characters more effectively and the recognition rate is higher than that of the traditional algorithm in the same conditions.

[26] made a comparative study involving four different feature extraction techniques for fingerprint classification. Also, they proposed a rank level fusion scheme for improving classification performance. They compared two well-known feature extraction methods based on Orientation Maps (OMs) and Gabor filters with two other new methods based on minutiae maps and orientation collinearity. Each feature extraction method was compared together in terms of accuracy and time. Moreover, they investigated on improving the classification performance using rank-level fusion. During the evaluation of each feature extraction method, their experimental results show that OMs performed best, in which Gabor feature fell behind OMs mainly because their computation is sensitive to errors in localizing the registration point. When fusing the rankings of different classifiers, they found that combinations involving OMs improve performance. Generally, the best classification results were obtained when they fused orientation map with orientation collinearity classifiers.

### III. PROPOSED BIMODAL FEATURE EXTRACTION FUSION

The proposed idea gives rise to an innovative way to fuse the features of two different modalities/traits. In the case of this paper face and fingerprint images were used. The procedure of the proposed bimodal fusion is grouped in the following basic stages:

1. Preprocessing stage, which involves the use of histogram equalization in enhancing the image(s) to be recognized.
2. Features of each modality on the preprocessed image are extracted using Gabor filter.
3. Dimensionality reduction using principal component analysis to reduce the dimension of the feature vectors due to their high dimensionality.
4. The fusion stage combined the corresponding feature vectors of each modality. The features are fused using the Mahalanobis distance. These distances are normalized by applying hyperbolic and fused using the average sum rule tanh [25].

#### A. Histogram equalization

An image histogram is a graphical representation of the tonal distribution in a digital image [11]. When viewing an image represented by a histogram, what really happens is analyzing the number of pixels vertically with a certain frequency horizontally. In essence, an equalized image is represented by an equalized histogram where the number of pixels is spread evenly over the available frequencies. These respective areas of the image that first had little fluctuation will appear grainy and rigid, thereby revealing other unseen details.

In order to equalize the face and fingerprint image histogram the cumulative distribution function (cdf) has been computed. The cdf of each gray level is the sum of its recurrence and the recurrence of the previous gray level in the image. The histogram equalization equation is given as:

$$h(v) = round\left(\frac{cdf(v)-cdfmin}{(W \times H)-cdfmin} \times (N-1)\right) \quad (3)$$

where $cdf_{min}$ is the minimum value of the cumulative distribution function, W and H are the width and height of image, N is the number of gray levels used. This result is an equalized and normalized image.

#### 1. Histogram equalization for fingerprint image enhancement

Fingerprint images are not 100 percent perfect, they may be affected by noise due to some factors, including: noise on capturing devices, the tip of the finger from which the measurements are taken and ridge patterns may be affected by cuts, dirt, or wear and tear. Also, it was noted that most of the fingerprint feature extraction is carried out using minutiae-based method and image-based method, which at times require an extensive preprocessing operation which includes normalization, segmentation, orientation, ridge filtering, binarization, and thinning. With all these stages of preprocessing there is degradation in the image to be extracted, whereby reducing the salient features [5][25]. In our paper histogram equalization was adopted, which enhanced the contrast of the fingerprint image (increasing the quality of the image). Histogram equalization is capable of revealing unseen details in an image. The result of the enhanced contract image is shown in Figure 2. Which produced an enhanced fingerprint image that is suitable in feature extraction.

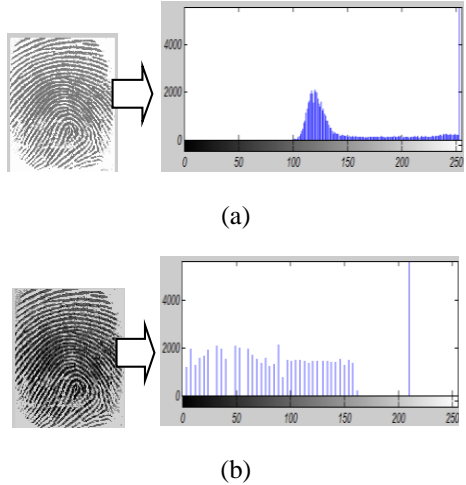

(a)

(b)

Figure 2: Fingerprint Image (a) before enhancement (b) After enhancement (histogram equalization)

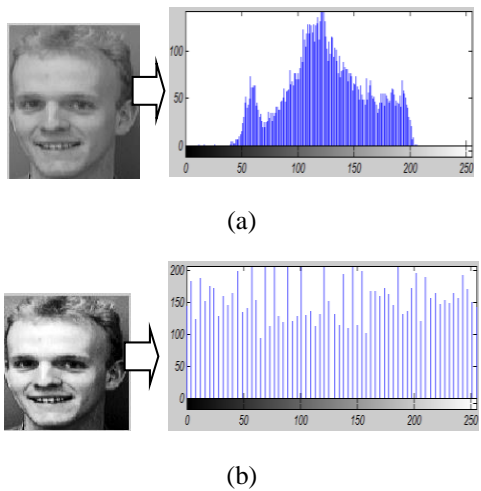

(a)

(b)

Figure 3: Face Image (a) before enhancement (b) After enhancement (histogram equalization)

*2. Histogram equalization for face image enhancement*

Face is considered to be the most commonly used biometric trait of humans, since it has shown its importance over the last ten years or so. Not only is it an intensely researched area of image analysis, pattern recognition and even biometrics to be precise [18] but it has also become an important in our daily lives since it was introduced as one of the methods for identification to be used in e-passports [18]; we recognize each other and in many cases establish our identities based on faces. Some of the benefits of using face image as one of the bimodal traits in this paper are: it is not intrusive, can be done from a distance even without the users being aware they are being scanned, and can also be used for surveillance purposes (as in searching for wanted criminals, suspected terrorists, or missing children).

Face images are not 100 percent perfect as long as they can be affected by many factors such as environment condition (stress), age, pose, illumination, facial expression, as well as changes in appearance due to make-up, facial hair [2]. Some of these problems are solved through preprocessing of the face image. In this paper, histogram equalization was used in enhancing the image contrast, in order to reveal some of the unseen details in the face image. Figure 3 shows the result of the enhanced contract image produced which is suitable for feature extraction.

*B. Gabor filter*

Gabor filter is a linear filter which is used for edge detection. Its frequency and orientation are similar to that of the human visual system, and they have been found to be appropriate for texture discrimination and representation. Gabor filters are formed by modulating a complex sinusoid by a Gaussian function. Gabor filters have been used widely in pattern analysis application, and it has been proved in extracting more salient features both in the face [1] and fingerprint [6][14][26] images, which are the two modalities being used in this paper. A set of Gabor filters with different frequencies and orientations was used for extracting salient features from both face and fingerprint images. It is invariant against translation, rotation, and variations due to illumination and scale. Gabor filter also presents desirable characteristics of spatial locality and orientation selectivity. During feature extraction the dimension or size of the image does not change. For instance, in this paper the dimension of both face and fingerprint is 112x92, after applying Gabor filter to extract the salient features the dimension still remains the same. This motivated the use of the principal component analysis (PCA) as a dimensionality reduction technique.

Each filter of Gabor is defined by:

$$G_{\lambda,\theta,\varphi,\sigma,\gamma}(x,y) = \exp\left(-\frac{x'^2 - \gamma y'^2}{2\sigma^2}\right)\cos\left(2\pi\frac{x^2}{\lambda} + \varphi\right)$$

where:

$$x' = x\cos\theta + y\sin\theta,\ y' = -x\sin\theta + y\cos\theta,$$

and $\lambda$, $\theta$, $\varphi$, $\sigma$ and $\gamma$ are wavelength, orientation(s), phase offset(s), bandwidth and aspect ratio respectively.

*1. Gabor filter techniques for extracting face and fingerprint features*

In this paper, the enhanced face and fingerprint images are used to extract the identifiable features in face and fingerprint. Figure 4 shows the performance of the new method when the number of orientations varies and the number of scales was fixed. The feature extraction information was based on the use of Gabor filtering parameters which includes: wavelength, orientation(s) degree, phase offset(s) aspect ratio, and bandwidth with their respective values.

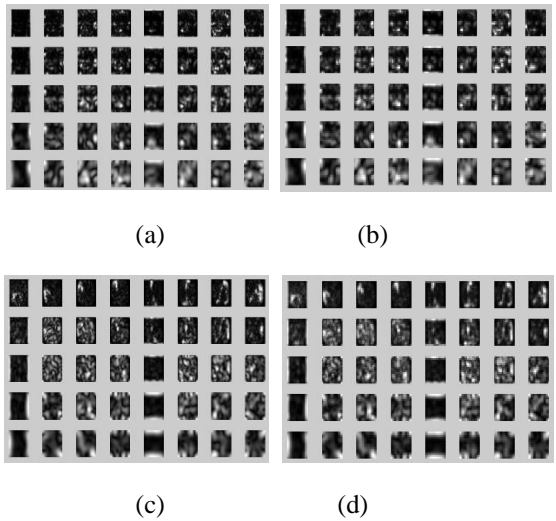

(a)　　　　　　(b)

(c)　　　　　　(d)

Figure 4: Gabor filter transformation of a sample of the face and fingerprint. (a) & (c) Magnitude responses of the filtering operation with the Gabor filter bank with no downsampling respectively (b) & (d) Magnitude responses of the filtering operation with the Gabor filter bank with downsampling 64 respectively.

### C. Principal component analysis (PCA)

The curse of dimensionality is a major problem in the extracted feature vector during the extraction of the salient features on both face and fingerprint images. This is as a result of higher dimensional space, which results in an enormous amount of data to be required to learn. PCA is one of the most commonly used dimensionality reduction techniques. It finds the principal components which is the underlying structure in the data. Figure 5 shows the performance of the PCA on both face and fingerprint by finding their PC. It helps in speeding up the algorithms and reduces space used by the data during the training, validation and testing stages using a supervised learning neural network (MLP) and also improves how we display information in a tractable manner for human consumption.

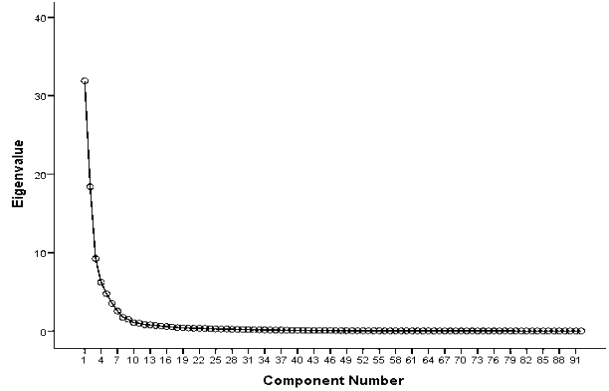

(a)

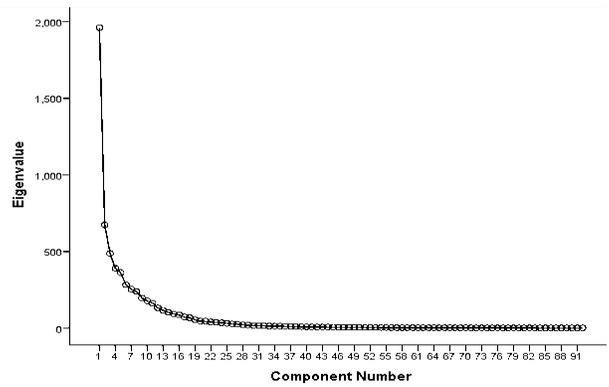

(b)

Figure 5: Principal component performance on (a) face (b) fingerprint.

### D. Mahalanobis distance technique

#### 1. Fusion of face and fingerprint feature vectors

[25] proposed a novel technique mahalanobis distance technique in the fusion of fingerprint and iris feature vectors at the feature extraction level. They compared this technique with other techniques and it was proved to be easier and more effective to use, due to the facts that other feature fusion performed serially or parallel, which at the end results in a high dimensional vector. But their proposed algorithm generates the same size fused vector as that of unimodal.

In this paper, the feature vectors extracted from the input images are combined together to form a new feature vector by making use of the adopted method feature level fusion technique. The extracted features from face and fingerprint are homogeneous, each vector are processed to produce the fused vector of the same dimension.

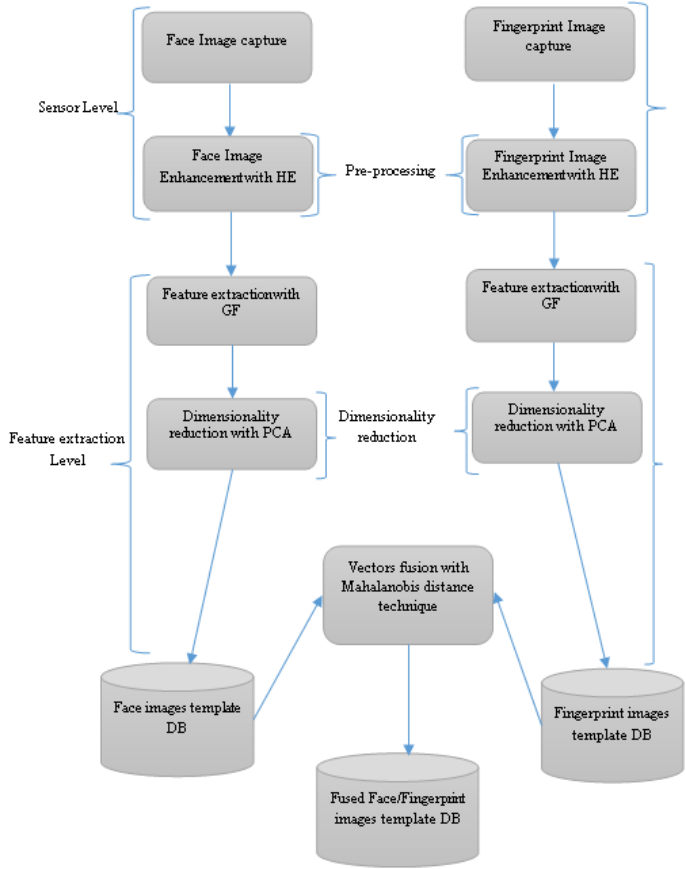

Figure 6: Architecture for the enhanced accuracy of biometric feature extraction fusion

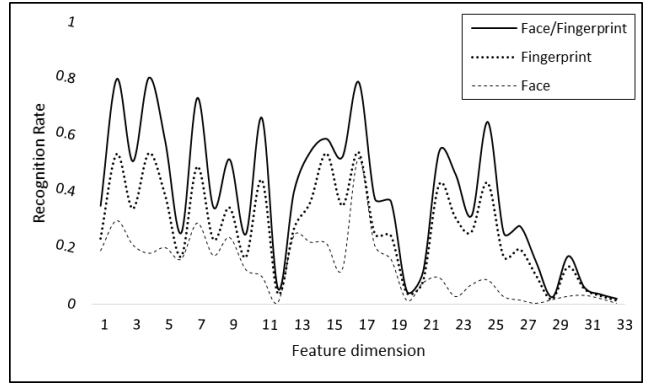

Figure 7: Recognition rate for different modalities

## IV. EXPERIMENTAL RESULTS

The performance evaluation of the proposed method is analyzed using the ORL face database and the ATVS fingerprint database. The ORL face database (http://www.face-rec.org/databases/) contains images from 40 individuals, each providing 10 images of different pose, expressions and decorations. The ATVS fingerprint database (http://atvs.ii.uam.es/databases.jsp) contains images from 17 individuals, each providing 4 different positions each of right middle and the small finger. Based on the numbers of fingerprint per each individuals, the experiments are performed using the first four images samples per class for testing the new approach.

### A. Comparison of face, fingerprint and the fusion

The results in Figure 7 show the performance rates of the extracted features for the face, fingerprint and the fusion of the two modalities. From the results, it was proven that the fusion of the two modalities have performance rate compared to the individual modalities.

### B. Comparison with other feature extraction methods

In this section the Gabor filter based feature extraction is compared to the KFA, LDA, PCA, and KPCA as shown in Table 1 for the face features, Table 2 for the comparison between Gabor filter and minutia based feature extraction for the fingerprint features, and Table 3 presents the results for the fusion of the face and fingerprint modalities using Gabor filter algorithm compared with other forms of feature extraction algorithms.

TABLE I: FEATURE EXTRACTION PERFORMANCE FOR FACE IMAGES

|  | KFA | LDA | PCA | KPCA | Gabor filter |
|---|---|---|---|---|---|
| **Face** | 82.35 % | 91.18 % | 97.01 % | 96.54 % | 97.35 % |

TABLE II: FEATURE EXTRACTION PERFORMANCE FOR FINGERPRINT IMAGES

|  | Minutia Based | Gabor filter |
|---|---|---|
| **Fingerprint** | 97.76% | 98.87% |

TABLE III: FEATURE EXTRACTION PERFORMANCE FOR THE FACE AND FINGERPRINT FUSION.

|  | KFA/Minutia Based | LDA/Minutia Based | PCA/Minutia Based | KPCA/Minutia Based | Gabor filter |
|---|---|---|---|---|---|
| **Face/fingerprint** | 90.01% | 94.47% | 97.39% | 97.15% | 98.11% |

When compared with what other researchers have done, the proposed method of fusing face and fingerprint modalities together using the Gabor filter technique for their feature extraction and Mahalanobis distance technique for fusing the two feature vectors together look promising.

V. CONCLUSION AND RECOMMENDATION

This paper presents an efficient way of extracting salient features in bimodal biometric systems. The proposed method uses face and fingerprint modalities from which the features were extracted. An image enhancement histogram equalization technique is used to enhance the face and fingerprint images. Salient features of the face and fingerprint were extracted using the Gabor filter technique. A dimensionality reduction technique is carried out on both images extracted features using a principal component analysis technique. A feature level fusion algorithm is used to combine each unimodal feature using the Mahalanobis distance technique. The performance of the proposed approach is validated and compared with other methods.

VI. FUTURE WORK

In order to fully enhance the accuracy and performance of the biometric recognition system, the proposed method can be trained and tested using a multilayer perceptron neural network which is a powerful non-linear classifier, produces elegant solutions built of continuous basic functions, has the ability to handle noisy data, and is fast to run.

ACKNOWLEDGMENT